\documentclass{INTERSPEECH2023}

 \interspeechcameraready

\title{A Textless Metric for Speech-to-Speech Comparison}
\name{Laurent Besacier$^1$, Swen Ribeiro$^2$, Olivier Galibert$^2$, Ioan Calapodescu$^1$}
\address{
  $^1$Naver Labs Europe, France\\
  $^2$LNE, France}
\email{laurent.besacier@naverlabs.com}

\begin{document}

\maketitle
 
\begin{abstract}
%This paper proposes a textless speech-to-speech comparison metric that allows comparing a speech hypothesis with a speech reference without falling-back to their text transcripts. We leverage recently proposed speech2unit encoders (such as HuBERT) to pseudo-transcribe the speech utterances into discrete acoustic units and propose a simple neural architecture that learns a speech-based metric which correlates well with its text-based counterpart. Such a textless metric could ultimately be interesting for speech-to-speech translation evaluation: for oral languages, for languages with no reliable ASR system available or simply to avoid backoff to ASR transcription.
In this paper, we introduce a new and simple method for comparing speech utterances without relying on text transcripts. Our  speech-to-speech comparison metric utilizes state-of-the-art speech2unit encoders like HuBERT to convert speech utterances into discrete acoustic units. We then propose a simple and easily replicable neural architecture that learns a speech-based metric that closely corresponds to its text-based counterpart. This textless metric has numerous potential applications, including evaluating speech-to-speech translation for oral languages, languages without dependable ASR systems, or to avoid the need for ASR transcription altogether. 
\end{abstract}

\section{Introduction}

\begin{figure}[ht!]
\includegraphics[width=0.41\textwidth]{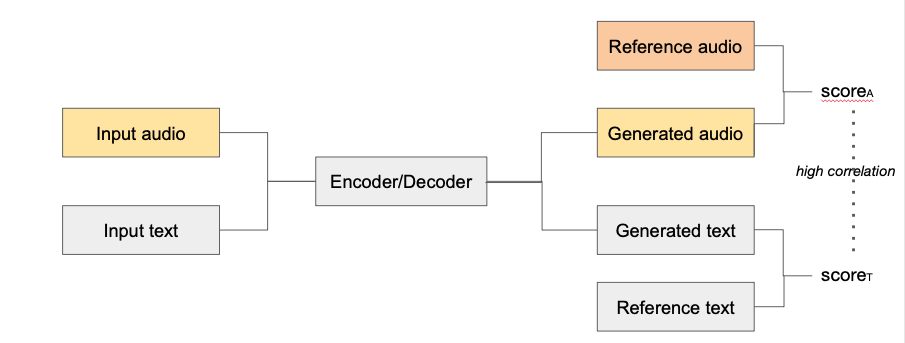} \newline
\caption{\label{goal}Our objective is to create a metric based on speech that strongly correlates with metrics based on text.}
\end{figure}

In natural language processing (NLP), matching a text hypothesis with a text reference is a common practice to evaluate systems such as natural language generation, machine translation, etc. With the rise of speech generation and end-to-end speech-to-speech (S2S) translation systems \cite{DBLP:journals/corr/abs-1904-06037,lee-etal-2022-direct}, there is a growing need for speech-to-speech comparison directly in the signal domain \cite{DBLP:journals/corr/abs-2110-13877}. This paper proposes a simple and efficient implementation of such "textless" metric.
More specifically, we want to develop a metric in order to compare a speech hypothesis ($H$) with a speech reference ($R$) along several axes. In this work, our main axis is \textit{meaning}, i.e similarity score should be high if both utterances convey same message. But other axes could be interesting in the future: eg. high similarity if $H$ and $R$ voices are similar (similar speaker, gender, etc.). We want our textless metric to have a strong correlation with its text-based counterpart that would be applied to the transcripts of $H$ and $R$ (see figure \ref{goal}). We believe  such  metric could be interesting for following use cases:  (a) evaluating a S2S translation system w/o falling back to a transcription of $H$ and $R$ (unlike sentence-level ASR-BLEU \cite{DBLP:journals/corr/abs-1904-06037}  does); (b) evaluating target languages for which we cannot fall back to a transcription such as Tamasheq \cite{DBLP:journals/corr/abs-2201-05051} ($>$50\% of languages are oral; even more are not equipped with good ASR) 
%or Hokkien;\footnote{\url{https://ai.facebook.com/blog/ai-translation-hokkien/}}
and (c) defining training objective for end-to-end S2S model optimization.

This paper is structured as follows: section \ref{background} positions our contribution in respect to previous works. Section \ref{rationale} describes a naive approach that fails and supports our proposal of a learnt metric. Section \ref{metric} presents our textless metric while section \ref{experiments} illustrates its use through speech comparison experiments with synthetic and natural speech. Finally section \ref{conclusion} concludes this work and gives some perspectives.

\section{Background}
\label{background}

\subsection{Text-to-Text Comparison Metrics}

String comparison in automatic speech recognition (ASR) evaluation often utilizes word error rate (WER). However, more specific metrics for machine translation evaluation such as BLEU \cite{BLEU} and TER (Translation Edit Rate) \cite{snover-etal-2006-study} have been proposed to go beyond simply counting insertions/deletions/substitutions. Later, ChrF (character-level F-score) \cite{popovic-2015-chrf} was proposed to address several shortcomings of BLEU such as sensitivity to tokenization. Finally since 2020 the use of contextualized text representations was experimented for evaluation such as in BERTScore \cite{DBLP:conf/iclr/ZhangKWWA20}.

\subsection{Learnt Metrics}

The common feature of metrics described in the previous paragraph is that they are  unsupervised: they compare two sequences of tokens and the hope is that the obtained scores will correlate well with human judgements (manually obtained quality scores). On the other hand, learnt metrics such as BEER \cite{BEER} and COMET \cite{COMET} are specifically trained to correlate with human judgments.
Actually, COMET\footnote{\url{https://unbabel.github.io/COMET/}} is more than a single metric and should rather be seen as an open-source framework for machine translation evaluation that can be used to: (a) evaluate systems with pre-trained metrics, and (b) train and develop new metrics. We will  adapt COMET to speech inputs in this work. 

\subsection{Speech Comparison and Textless NLP}

Speech-to-speech comparison is not an unexplored territory. The first approaches for isolated word recognition from speech used Dynamic Time Warping (DTW) \cite{sakoe1978dynamic} to measure a distance directly between two speech signals. DTW aligns two sequences of feature vectors by warping their time axes to achieve an optimal match. However, even if DTW is still used for word spotting applications, it is ill-equipped to reach our goal of measuring subtle differences in meaning between long speech hypothesis and reference.
These limitations of signal-based comparison metrics such as DTW lead us to get interested by the textless NLP area \cite{kharitonov-etal-2022-textless}. One building block of this emerging domain is the use of Speech-to-Units (S2U) encoders that automatically discover discrete acoustic units and decode speech into a pseudo-text. Examples of such encoders are HuBERT \cite{HuBERT} or Wav2Vec2.0 \cite{wav2vec2.0} followed by a quantization function (using k-means algorithm for instance). Such representations were successfully used for automatic speech recognition (ASR) tasks which shows that, whether discretized or not, they convey information related to text message hidden in  speech signal. 
%Therefore, our intuitive and initial attempt to build a textless metric is to use those S2U encoders to "transcribe" speech into a pseudo-text and then re-use well-known text-to-text comparison metrics.
%as described in the next section.

Contemporaneous to this work, \cite{https://doi.org/10.48550/arxiv.2212.08486} propose a text-free metric for S2S evaluation  but they train it on human annotations (to correlate with human judgements) whereas we will train our S2S evaluation metric to correlate with its text-based counterpart (which will allow us to take advantage of much more data as no human evaluation data is needed in our case). 
%leveraging all tand we show how our metric correlates with human annotations 

\section{A (Too) Naïve Approach}
\label{rationale}

Our first attempt was to apply text-based translation metrics to our pseudo-transcribed (S2U) speech signals.  Discrete acoustic units were generated after clustering audio features; standard machine translation metrics such as BLEU were then applied to the unit sequences obtained.  We  then verified if speech-BLEU would correlate with text-BLEU when applied to multiple pairs $(H,R)$ of utterances.
%same utterances, but on the true orthographic transcripts. 
We applied the following experimental setup: (a) build a dictionary of $k$ centroïds from large speech data using k-means algorithm applied to cepstral features; (b) pseudo-transcribe the pairs of speech utterances by mapping each feature vector to the nearest centroïd (using l2 or cosine distance); (c)  reduce consecutive repetitions of the same discrete symbol into one instance (de-duplication).

\begin{figure}[ht!]
\centering
\includegraphics[width=0.32\textwidth]{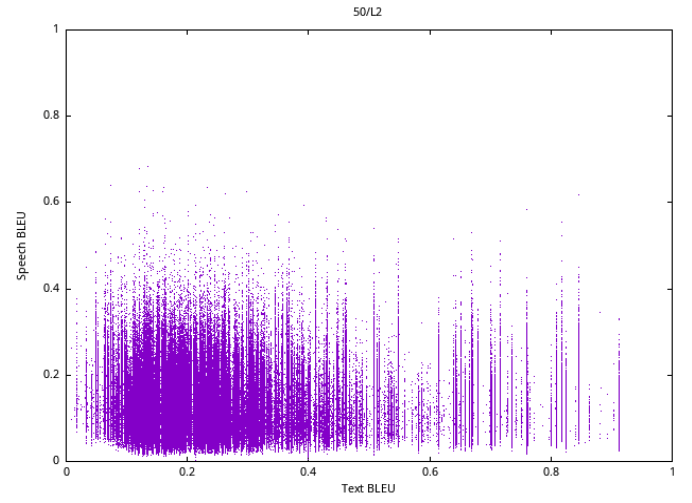}%
\caption{\label{failure}Scatterplots of text-BLEU (X axis) versus speech-BLEU (Y axis) for a vocabulary of 50 acoustic units and l2 distance to the centroïd. We observe no correlation between text-based and speech-based metric with this naïve approach.}
\end{figure}

We computed speech-based metrics (and their text-based counterpart) on a subset of commonvoice4.0 english dataset.\footnote{\url{https://commonvoice.mozilla.org/}} More precisely, we selected 20M pairs of utterances with at least one 4-gram in common (in order to have non-zero BLEU scores in our collection).
Figure \ref{failure} presents a scatterplot of the text-based metric (on X axis) and speech-based metric (on Y axis) for a subset of those pairs when a vocabulary of k=50 acoustic units is used (we experimented with different values of k and a cosine metric instead of l2 with similar results).
%can be found in Appendix \ref{sec:appendix1}).

Initially, we noted that our selection procedure, which involves choosing pairs of natural speech utterances that share at least one 4-gram, enables the collection of pairs with a wide range of BLEU scores between 0 and 1, reflecting their varying degrees of closeness. However, naive speech-BLEU does not correlate well with text-BLEU which shows that text-based metrics simply applied to discrete speech units fail. This leads us to propose a new approach that differs in two main points:
\begin{itemize}
    \item instead of local acoustic (cepstral) quantized units, we use HuBERT \cite{HuBERT} units that have been shown to convey more contextualized and semantic speech information,
    \item  simple unsupervised (such as BLEU) metrics used are replaced by learnt metrics.
    %(such as the ones proposed in COMET).
\end{itemize}

\begin{figure*}[!ht]
    \centering
    	\includegraphics[width=0.8\textwidth]{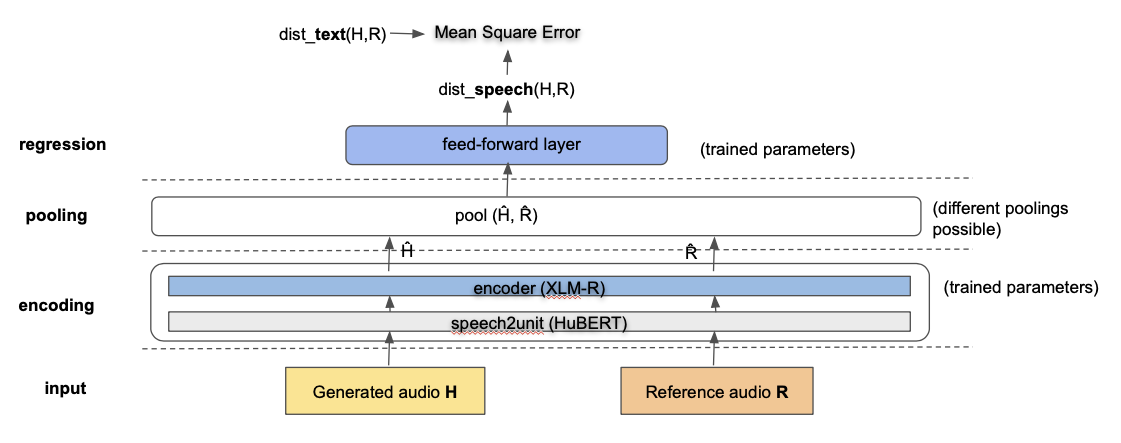}
        \caption{Adaptation of the COMET framework to our textless metric.}
    	\label{fig:comet-speech}
\end{figure*}

\section{A Learnt Metric for Speech-to-Speech Comparison}
\label{metric}

As it has been observed that text-based metrics applied to discretized acoustic speech units are unreliable, we believe that the best approach is to develop a metric that learns the semantic similarity between a speech hypothesis and a speech reference.
To experiment with this idea, we re-use COMET  \cite{COMET} framework widely adopted in machine translation evaluation where it is trained to correlate with human judgements. We  adapt it  to our need as illustrated in figure \ref{fig:comet-speech}: both audio $H$ and $R$ are pseudo-transcribed in a sequence of de-duplicated speech units (with HuBERT \cite{HuBERT}). Both sequences of discrete units are mapped to a sequence of characters\footnote{Each discrete unit is mapped to a rare character in the Unicode set} and encoded with a neural text encoder. Obtained $\widehat{H}$ and $\widehat{R}$ vectors are pooled and a regression layer predicts the score we want to approximate. 

In the follow-up experiments we use ChrF or BLEU text-based metrics as a target. Mean Square Error (MSE) loss is used to train model parameters. As done in initial COMET framework, not only regression layer parameters will be learnt during training but also parameters of the "text" encoder (after 30\% of the first epoch and for the rest of the training steps).
We highlight below  main differences between initial COMET and its adaptation to speech-to-speech comparison:
\begin{itemize}
    \item  COMET uses source $S$, hypothesis $H$ and reference $R$ utterances to predict MT quality, whereas we only use $H$ and $R$ in this work ($S$ is ignored during pooling operation),
    \item  COMET predicts human judgement of MT quality whereas our learnt metric predicts a text-based score (no human judgements are needed to train our metric),
    \item COMET proposes two different training objectives: \textit{regression} to predict a score or \textit{ranking} using a triplet loss; 
    %to encode good translations closer to the reference while pushing bad translations away; 
    we use only \textit{regression} here,
    \item COMET comes with several text encoders (XLM-Roberta \cite{xlm-roberta}, BERT \cite{devlin-etal-2019-bert}); we use XLM-Roberta (277M param.) to encode our sequence of discrete acoustic units as we believe it should be able to capture sequential patterns; XLM-Roberta parameters are fine-tuned after 30\% of first epoch.\footnote{Using a true speech encoder such as XLSR \cite{XLSR} to replace the stacking of HuBERT and XLM-Roberta is an option left for future work as it would require major modification of COMET codebase.
    %and might not even be needed as we will see that stacking HuBERT and XLM-R already gives strong results.
    }
\end{itemize}

\section{Experiments}
\label{experiments}

In order to show that our approach can learn several metrics,
we first experiment on English synthetic speech and train a metric to predict speech-ChrF. In a second step we predict speech-BLEU using English natural (human) speech.

\subsection{ChrF prediction on synthetic speech}

\begin{figure}[!ht]
    	\includegraphics[scale=0.2]{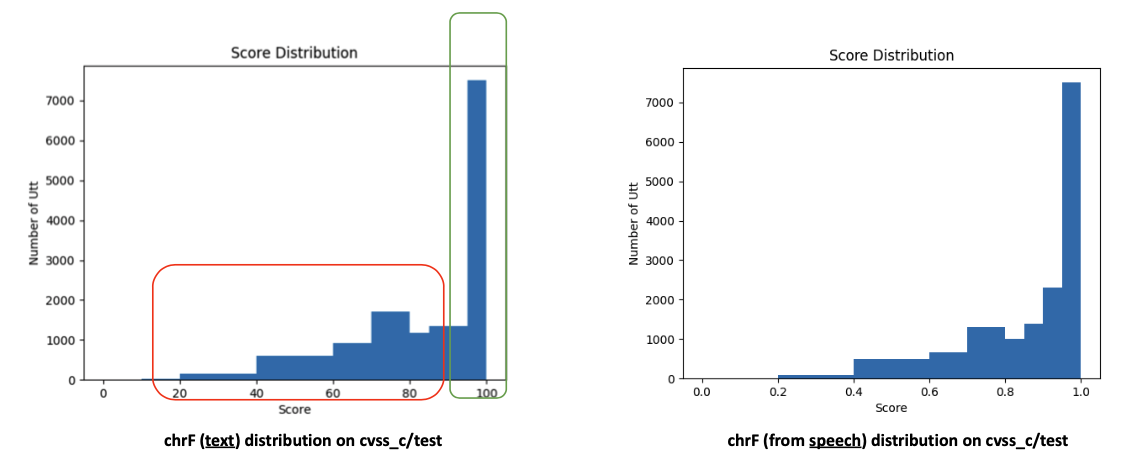}
    	\caption{Distributions of ChrF scores on the test set of  prepared CVSS corpus: (left) text-ChrF (right) speech-ChrF.}
    	\label{fig:distribs}
\end{figure}

 We start from synthetic CVSS speech corpus \cite{cvss-corpus}  (English target part), a massively multilingual-to-English S2S corpus. To obtain dissimilar audios with different voices, we enrich CVSS using the following process applied to each English speech utterance: (a) ASR transcription; 
 %using SpeechBrain toolkit \cite{speechbrain}; 
 (b) BART \cite{lewis-etal-2020-bart} encoding and decoding to further add noise to the already noisy ASR transcript; and (c) TTS from the noisy transcript (with a different speaker voice).
 We end-up with a corpus of 256,882 pairs ($H$,$R$) of speech utterances (similar, slightly dissimilar or very dissimilar) with associated transcripts  splitted into train (207,364 pairs), dev (14,759 pairs) and test (14,759 pairs). 
 %Figure \ref{fig:distribs} (left) shows the distribution of our test set according to the true (text-based) ChrF score.
 %All $H$ and $R$ are synthetic speeches from different speaker voices. They correspond to similar or (slightly) dissimilar audios as illustrated by the 
 True text-ChrF distribution (on our test set) is displayed in figure \ref{fig:distribs}  (left).
 %obtained from their text transcripts. 
We learn several metrics using COMET and display correlations (Pearson and Spearman) between true ChrF and learnt ChrF for different setups (table \ref{tab:comet-results}):
\begin{itemize}
    \item different input (text or speech),
    \item different amount of training data (for learnt metrics): \textit{dev} set (14.7k utterances) or \textit{train} set (207.4k utterances),
    \item different number of HuBERT acoustic units: 50 or 200,
    \item different  number of training epochs: 5 or 10.
\end{itemize}

First row in table \ref{tab:comet-results} is a topline were ChrF was learnt (using our \textit{dev} set) with initial COMET framework and \textit{text} inputs. As expected, neural architecture can learn to approximate a sequence based metric such as ChrF easily (high correlation between true and predicted ChrF scores).  Remaining rows use speech $H$ and $R$ inputs: second row is the naive baseline presented in section \ref{rationale} with poor correlation scores. Rows 3-6 display results obtained with our learnt textless metric (speech-ChrF). We observe that more acoustic units (200 instead of 50), adding training data (207k utterances instead of 14.7k utterances) and training longer (10 epochs instead of 5)  improves correlation.
To illustrate better what 0.779 Pearson correlation score means, figure \ref{fig:distribs} (right) displays distribution of our speech-ChrF scores (with best configuration of last row in  table \ref{tab:comet-results}). We observe that left (text-ChrF) and right (speech-ChrF)  distributions are very similar. Our modified COMET has learnt to replicate the text-ChrF distribution using speech input only.

\begin{table*}[]
\centering
\caption{Correlations between text-ChrF and speech-ChrF  (on synthetic speech, test set) for different experimental setups.}
\label{tab:comet-results}
{\footnotesize
\begin{tabular}{|l|l|l|l|l|l|l|}
\hline
Input & Train Data & Encoder                                           & Epochs & Metric & $\rho$ (Pearson) & $\rho$ (Spearman) \\ \hline
%\rowcolor[HTML]{FFFFC7} 
%Text           & None                & None                                                        & -               & chrF            & 1                    & 1                     \\ \hline
%\rowcolor[HTML]{FFFFC7} 
Text (topline)           & 14.7k utt.          & XLM-R                                                       & 5               & learnt chrF     & \textbf{0.902}                & \textbf{0.922}                 \\ \hline
%\rowcolor[HTML]{DAE8FC} 
Speech (baseline)         & None                & Hubert-50                                                   & -               & naive chrF            & 0.431                & 0.386                 \\ \hline
%\rowcolor[HTML]{DAE8FC} 
Speech          & dev (14.7k)          & \begin{tabular}[c]{@{}l@{}}Hubert-50\\ +XLM-R\end{tabular}  & 5               & learnt chrF     & 0.542                & 0.480                 \\ \hline
%\rowcolor[HTML]{DAE8FC} 
Speech          & dev (14.7k)          & \begin{tabular}[c]{@{}l@{}}Hubert-200\\ +XLM-R\end{tabular} & 5               & learnt chrF     & 0.595                & 0.567                 \\ \hline
%\rowcolor[HTML]{DAE8FC} 
Speech          & train (207.4k)        & \begin{tabular}[c]{@{}l@{}}Hubert-200\\ +XLM-R\end{tabular} & 5               & learnt chrF     & 0.755                & 0.700                 \\ \hline
%\rowcolor[HTML]{DAE8FC} 
Speech          & train (207.4k)       & \begin{tabular}[c]{@{}l@{}}Hubert-200\\ +XLM-R\end{tabular} & 10              & learnt chrF     & \textbf{0.779}       & \textbf{0.733}        \\ \hline
\end{tabular}
}%/medium
\end{table*}

%%%%Do the table
\begin{table*}[]
\centering
\caption{Correlations between (a) text-BLEU and speech-BLEU (b) text-BLEU and sentence-level ASR-BLEU (natural speech)}
\label{tab:natural}
{\footnotesize
\begin{tabular}{lllllll}
\hline
\multicolumn{1}{|l|}{metric} & speech-BLEU                         & \multicolumn{1}{l|}{(ours)}         & ASR-BLEU \cite{DBLP:journals/corr/abs-1904-06037}                         & \multicolumn{1}{l|}{(whisper tiny 28.8\% WER)} & ASR-BLEU \cite{DBLP:journals/corr/abs-1904-06037}                         & \multicolumn{1}{l|}{(whisper large 10.1\% WER)} \\ \cline{2-7} 
\multicolumn{1}{|l|}{}       & \multicolumn{1}{l|}{$\rho$ (Spearman)}    & \multicolumn{1}{l|}{$\rho$ (Pearson)}     & \multicolumn{1}{l|}{$\rho$ (Spearman)} & \multicolumn{1}{l|}{$\rho$ (Pearson)}     & \multicolumn{1}{l|}{$\rho$ (Spearman)} & \multicolumn{1}{l|}{$\rho$ (Pearson)}      \\ \hline
\multicolumn{1}{|l|}{dev}    & \multicolumn{1}{l|}{\textbf{0.838}} & \multicolumn{1}{l|}{\textbf{0.988}} & \multicolumn{1}{l|}{0.531}       & \multicolumn{1}{l|}{0.528}          & \multicolumn{1}{l|}{0.784}       & \multicolumn{1}{l|}{0.822}           \\ \hline
\multicolumn{1}{|l|}{test}   & \multicolumn{1}{l|}{\textbf{0.881}} & \multicolumn{1}{l|}{\textbf{0.976}} & \multicolumn{1}{l|}{0.579}       & \multicolumn{1}{l|}{0.593}          & \multicolumn{1}{l|}{0.771}       & \multicolumn{1}{l|}{0.805}           \\ \hline
                             &                                     &                                     &                                  &                                     &                                  &                                     
\end{tabular}
}%{\footnotesize
\end{table*}

\subsection{BLEU prediction on natural speech}

\subsubsection{Setup}

\begin{figure}[!ht]
        \centering
    	\includegraphics[scale=0.23]{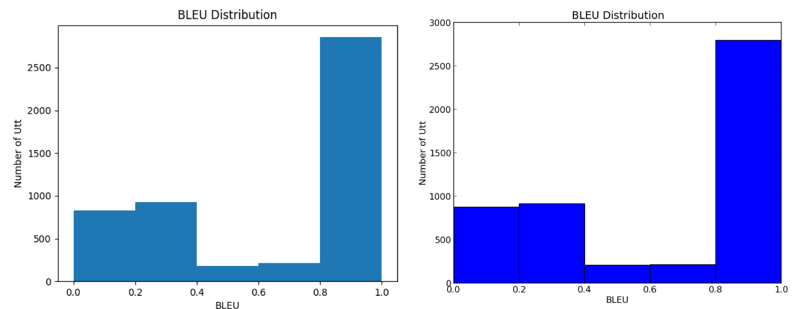}
    	\caption{Distributions of BLEU on natural speech (test): (left) text-BLEU; (right) speech-BLEU; train/dev have close distrib.}
    	\label{fig:natural}
\end{figure}

We now evaluate on natural speech. We use 1M pairs obtained with methodology described in section \ref{rationale} (commonvoice corpus) where $H$ and $R$ are natural speech utterances most of the time from different speakers. Target score is now BLEU metric obtained from text. Our corpus is splitted into \textit{train} (990k pairs), \textit{dev} (5k pairs) and \textit{test} (5k pairs). Figure \ref{fig:natural} (left) displays BLEU distribution of our test set (train/dev distributions are similar): distribution is bimodal with many unmatched pairs in range [0;0.4] and even more matched pairs in range [0.8;1.0]. 

\begin{figure}[!ht]
        \centering
    	\includegraphics[scale=0.23]{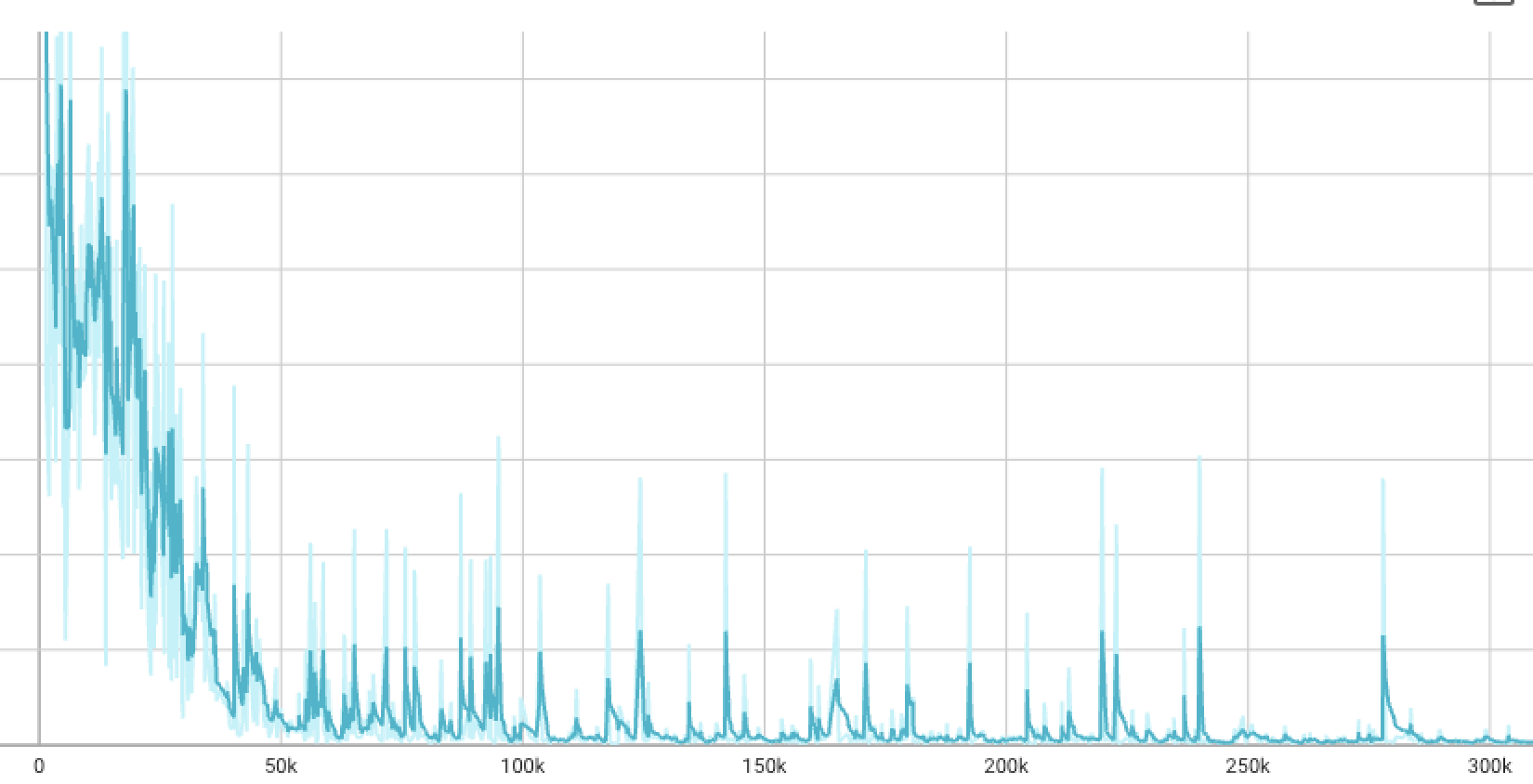}
    	\caption{Loss (MSE) during 5 epochs of training  speech-BLEU on 990,000 pairs of natural speech utterances.}
    	\label{fig:comet-natural}
\end{figure}

After extracting HuBERT-200 acoustic units for the full speech collection, we learn our  speech-BLEU on the \textit{train} set of 990,000 pairs of speech utterances (for 5 epochs) and evaluate on our \textit{dev} and \textit{test} (5k pairs each). Overall our model has 279M trainable parameters (among them 277M for XLM-R) and was learnt in 60h on a single GPU-V100. Training loss is displayed on figure \ref{fig:comet-natural}: we clearly see when parameters of the XLM-R encoder (after 20k steps corresponding to 30\% of the first epoch) start to be adapted in addition to the regression layer parameters. At this moment XLM-R specializes itself at encoding acoustic HuBERT units and loss significantly decreases.

We obtain very good correlations on the test set (see table \ref{tab:natural}): $\rho(Pearson)=0.976$ and $\rho(Spearman)=0.881$. This demonstrates that our approach, when learnt on enough pairs of natural speech, can be used to train a similarity metric (such as BLEU) between two audio speech samples.
The high correlation coefficients (actually higher than the ones obtained on synthetic speech) can be explained by the fact that our training data is bigger for natural speech (1M pairs) than for synthetic speech (207k pairs) and also by the BLEU distribution of our dataset (figure \ref{fig:natural}) which is bimodal with large majority of high scores ([0.8,1.0]) making  score prediction task probably easier.

\subsubsection{Comparison with sentence-level ASR-BLEU}
We compare our approach with sentence-level ASR-BLEU \cite{DBLP:journals/corr/abs-1904-06037} which consists in automatically transcribing both speech hypothesis and reference and compute sentence-level BLEU between  transcripts.\footnote{We however apply ASR on both $H$ and $R$ while \cite{DBLP:journals/corr/abs-1904-06037} applies it to $H$ only and use pre-existing text reference for $R$.} We used two multilingual ASR \cite{https://doi.org/10.48550/arxiv.2212.04356} models with different performance (whisper-tiny/39M parameters and whisper-large/1550M parameters) to decode signals of  \textit{dev} and \textit{test} sets (we keep punctuation and case for ASR-BLEU computation). Right part of table \ref{tab:natural} shows that sentence-level ASR-BLEU is a poor proxy to real text-BLEU even when ASR system is strong (whisper-large obtains 10.1\% WER on common voice data according to \cite{https://doi.org/10.48550/arxiv.2212.04356}). When ASR system is weaker (whisper-tiny obtains 28.8\% WER on common voice data according to \cite{https://doi.org/10.48550/arxiv.2212.04356})   correlation scores are even worse. 
This indicates that in situations where the target language lacks a robust ASR system, relying solely on ASR-BLEU could be misleading. Conversely, our trained speech-BLEU metric exhibits the strongest correlation with the original text-BLEU.

\subsubsection{Qualitative analysis}

%\textbf{ADD ANALYSIS and EXAMPLES HERE}
Figure \ref{fig:natural} (right) displays speech-BLEU distribution of our test set which is similar to the original text-BLEU distribution which confirms, for natural speech, results of figure \ref{fig:distribs} already found for synthetic speech.\footnote{Supplementary material provides distrib of  ASR-BLEU scores and show those are  different from text-BLEU and speech-BLEU ones.} 
As supplementary multimedia material, we offer audio pairs with their text-BLEU, speech-BLEU, and ASR-BLEU scores for randomly selected utterances from the \textit{test} set. Through our observations, we found that our speech-BLEU metric is capable of predicting low scores for poorly related utterances, while also indicating a score close to 1 for similar utterances spoken by different speakers.

%\subsection{Cross-lingual generalization of the metric}

%todo ?

\section{Conclusion and Future Work}
\label{conclusion}

%We have presented our architecture for learning a textless metric for speech-to-speech comparison. Preliminary results show that there is enough information in a sequence of discrete HuBERT units to score semantic similarity between an audio hypothesis and an audio reference even if speaker voices differ. However, a proper learning of  the metric is needed to increase its correlation with text-based scores. Next step of this work is to check if the metric can be learnt on multilingual data and can generalize to new unseen languages as this is the case for COMET in machine-translation evaluation. Indeed, one goal is to use the textless metric for evaluation of speech-to-speech translation into oral languages, such as recently done in the low-resource speech translation shared task at IWSLT2022.
We have introduced our architecture for developing a text-free metric for speech-to-speech comparison. Our initial findings indicate that a sequence of discrete HuBERT units contains enough information to measure semantic similarity between an audio hypothesis and reference, even when the speakers are different. However, proper training of the metric is necessary to improve its correlation with text-based scores. Moving forward, we aim to determine if the metric can be trained on multilingual data and applied to unseen languages, similar to COMET in machine translation evaluation. Another objective is to use the text-free metric to evaluate speech-to-speech translation into oral languages (for the low-resource speech translation shared task at IWSLT2022\footnote{\url{https://iwslt.org/2022/low-resource}}). 
%Finally, our metric could also be extended to other evaluation tasks: MOS prediction task in TTS which was recently proposed as a challenge of last Interspeech 2022\footnote{\url{https://voicemos-challenge-2022.github.io}} or ASR confidence measures (by simply replacing reference $R$ by source $S$ in our architecture).
Additionally, our metric can be applied to other evaluation tasks, such as the MOS prediction task in TTS, which was recently presented as a challenge at Interspeech 2022.\footnote{\url{https://voicemos-challenge-2022.github.io}} We could also adapt our architecture to ASR confidence measures by replacing the reference $R$ with the source $S$.

%\section{Acknowledgements}

%Part of the work presented here was carried out during the 2022 Jelinek Memorial Summer Workshop on Speech and Language Technologies at Johns Hopkins University (JSALT) within the group \textit{Speech Translation for Under-Resourced Languages}. Moreover, this work was performed using HPC resources from GENCI–IDRIS (Grant 2022-AD011012565 )

\bibliographystyle{IEEEtran}
%\bibliography{mybib}
\bibliography{anthology,custom,references}

\end{document}